# COMPARING PERFORMANCE OF DIFFERENT LINGUISTICALLY-BACKED WORD EMBEDDINGS FOR CYBERBULLYING DETECTION

Juuso ERONEN[1], Michal PTASZYNSKI[2] and Fumito MASUI[3]

[1]Doctoral Course in Manufacturing Engineering, Kitami Inst. of Tech.
E-mail: eronen.juuso@gmail.com
[2]Associate Professor, Dept. of Computer Science, Kitami Inst. of Tech.
E-mail: michal@mail.kitami-it.ac.jp
[3]Professor, Dept. of Computer Science, Kitami Inst. of Tech.
E-mail: f-masui@mail.kitami-it.ac.jp

(Koencho 165, Kitami, Hokkaido 090-8507, Japan)

In most cases, word embeddings are learned only from raw tokens or in some cases, lemmas. This includes pre-trained language models like BERT. To investigate on the potential of capturing deeper relations between lexical items and structures and to filter out redundant information, we propose to preserve the morphological, syntactic and other types of linguistic information by combining them with the raw tokens or lemmas. This means, for example, including parts-of-speech or dependency information within the used lexical features. The word embeddings can then be trained on the combinations instead of just raw tokens. It is also possible to later apply this method to the pre-training of huge language models and possibly enhance their performance. This would aid in tackling problems which are more sophisticated from the point of view of linguistic representation, such as detection of cyberbullying.

## 1. INTRODUCTION

The use of word embeddings in the representation of contextual information is a central concept in natural language processing. Word embeddings encode the meaning in a way that the words that are closer in the vector space are expected to be more similar. In the recent years word embeddings trained with neural networks have become a widely-used standard NLP technology [1, 2], and have been successfully applied to a variety tasks, including automatic cyberbullying detection [3, 4, 5].

In almost all cases, word embeddings are learned only from raw tokens (words) or in some cases, lemmas (unconjugated forms of words). This also applies to the recently popularized pre-trained language models like BERT [6]. Although word embeddings themselves have been used to predict linguistic information like parts-of-speech (POS) [7], named entity recognition (NER) [8], or dependency parsing [9], utilizing them in the training process of the embeddings themselves has not been extensively researched so far, with only a small number of related studies being proposed at this time [10, 11, 12, 13].

To further explore the potential of using linguistic information in the training process of word embeddings in order to capturing deeper relations between lexical items and structures and to filter out redundant information, we propose to preserve the morphological, syntactic and other types of linguistic information by combining them with the raw tokens or lemmas. This means, for example, explicitly including parts-of-speech or dependency information in the training of the embeddings. This means that the word embeddings can be trained using these features in combination with raw tokens. This method could also later be applied to the pre-training process of huge language models with the aim of enhancing their performance.

The structure of the paper is as follows. Firstly, we introduce the existing research in using linguistic information in training word embeddings and how it could show improvements in the field of automatic cyberbullying detection. Next, we describe how we utilized linguistic information in this study and introduce the classifiers used to evaluate the methods. Lastly, we introduce our experiment setup and discuss the effect of using linguistic information in the training of word embeddings.

## 2. RELATED WORK

**(1) Linguistic information in word embeddings**

There are only a handful of studies related to the usage of linguistic information in training word embeddings. In 2014, Levy and Goldberg [10] modified the Skip-Gram model used by Word2Vec [1] to use dependency structures as contexts while training the word vectors instead of using only a fixed window of surrounding words. They noticed that their dependency-based embeddings were noticeably different from the ones trained with words as contexts as they seemed to be more functional instead of topical. Their method was later evaluated by Komninos and Manandhar [11] and MacAvaney and Zeldes [14]. They acknowledged that dependency-based embeddings outperform the use of linear context in many tasks, especially question classification and semantic relation identification.

In 2017 Ptaszynski et al.[12] proposed a method of adding linguistic information like POS, NER and dependency structures, to the creation of bag-of-words (BoW) models. This showed an improvement to ordinary BoWs when using a Convolutional Neural Network (CNN) model. Their study also hinted that increasing (or decreasing) the density of the used feature set could result in an increased performance. In 2019, Cottorell and Schutze [13] proposed a method of keeping morphological information, like POS, case, gender etc. to encode the words' morphology. They showed that it is possible to encode such information better than Word2Vec by using a modified Log-Bilinear model [15].

**(2) Cyberbullying detection**

The research by Ptaszynski et al. [12] showed on actual cyberbullying data that using linguistically-backed preprocessing methods can be used to achieve higher classification performance. They noticed that a BoW model with encoded dependency information showed an improved performance when utilized with Convolutional Neural Networks. The reason behind this could be that cyberbullying, which, being a serious social problem, is a very sophisticated problem from the point of view of linguistic representation.

Other recent research in cyberbullying detection has mainly concentrated on using recurrent neural networks and pretrained language models with raw tokens to train embeddings [3, 4, 5]. The exceptions being Balakrishnan et al. [16] and Rosa et al. [17], who used psychological features, like personalities, sentiments and emotions to improve automatic cyberbullying detection. However, these were done using simple models. Using linguistic preprocessing and linguistic embeddings to improve classifier performance has not been studied further with cyberbullying detection even though the potential was confirmed earlier [12].

## 3. METHODS

We ran our experiments on the Kaggle Formspring Dataset for Cyberbullying Detection [18]. However, the original dataset had a problem of being annotated by laypeople, whereas it has been pointed out before that datasets for topics such as online harassment and cyberbullying should be annotated by experts [19]. Therefore in our research we applied a version re-annotated with the help of highly trained data annotators with sufficient psychological background to assure high quality of annotations [20]. The dataset contains almost thirteen thousand expert annotated samples. The number of harmful samples is small, amounting to 7% of the total samples. However, this roughly reflects the typical amount of profanity on SNS [19].

In this study, we trained Word2Vec Skip-Gram embeddings with encoded linguistic information and also by using dependency structure based contexts similarly to Levy and Goldberg [10]. We then evaluated these methods using different kinds of neural network models with Support Vector Machines [21] as the baseline.

**(1) Linguistically-backed word embeddings**

In order to train the linguistically-backed embeddings, we first preprocessed the dataset in various ways, similarly to Levy and Goldberg [10] and Ptaszynski et al. [12]. The preprocessing was done using spaCy NLP toolkit (https://spacy.io/).

- **Tokenization:** words separated by spaces (later: TOK).
- **Lemmatization:** like the above but in generic (dictionary) forms of words ("lemmas") (later: LEM).
- **Encoded parts of speech:** parts of speech information is merged with LEM or TOK (later: POS).
- **Encoded dependency structures:** token pairs with syntactic relations encoded between them (later: DEP).
- **Dependency-based contexts:** The use of dependency relations instead of a fixed window of tokens as context when training embeddings (later: DEPC) [10].

We generated a Word2Vec Skip-Gram language model from each of the processed dataset versions. This resulted in separate models for each of the datasets, Tokens-Skip-Grams, Lemmas-Skip-Grams, Tokens-POS-Skip-Grams, Lemmas-POS-Skip-Grams, DEP-Skip-Grams and DEPC-Skip-Grams.

The dependency-based context embeddings (DEPC-Skip-Grams) are the same dependency embeddings used in the previous research by Levy and Goldberg [10]. Other embeddings used a fixed context window size of 5. The embeddings were pretrained on a 1GB sample of English Wikipedia dataset using Gensim [22] with 300 dimensions.

**(2) Classification**

To evaluate the embeddings, we used a linear Support Vector Machine (SVM) [21] as the baseline. We also used different neural network architectures, Recurrent Neural Network with Long short-term memory (LSTM), Convolutional Neural Network (CNN) and Multilayer Perceptron (MLP).

As the baseline, we applied SVMs [21], which are a set of classifiers well established in Artificial Intelligence and Natural Language Processing. SVM also has had much success in previous cyberbullying research [23].

We applied an LSTM implementation with Hyperbolic Tangent (tanh) as a neuron activation function, and dropout regularization. We used Adaptive Moment Estimation (Adam), a variant of Stochastic Gradient Descent [24] as the optimizer.

We applied a CNN implementation with Rectified Linear Units (ReLU) [25] as a neuron activation function, and max pooling [26], which applies a max filter to non-overlying subparts of the input to reduce dimensionality and in effect correct overfitting. We also applied dropout regularization on penultimate layer, 4x4 size of patch and 2x2 max-pooling.

In this experiment MLP refers to a network using regular dense layers. We applied an MLP implementation with Rectified Linear Units (ReLU) as a neuron activation function and one hidden layer with dropout regularization which reduces overfitting and improves generalization by randomly dropping out some of the hidden units during training [25]. The neural network models used in this study were trained using Keras [27].

## 4. EXPERIMENTS

**(1) Setup**

The preprocessing provides 7 separate datasets for both the Wikipedia dataset and the target cyberbullying dataset. We trained the embeddings and performed the experiments once for each type of preprocessed dataset. Each of the classifiers (sect. (2)) were tested on each version of the dataset in a 10-fold cross validation procedure. The evaluation results were calculated using balanced F-score. As the dataset was not balanced, we weighted the classes accordingly. We ran two sets of experiments. First pretraining the embeddings on the Wikipedia dataset prior to training the classifiers on the target cyberbullying dataset. Second, we trained the embeddings *ad hoc* on the target dataset itself.

**(2) Evaluation of linguistic embeddings**

From the results presented in the upper half of Table 1 it can be seen that most of the classifiers scored highest on raw lemmas embeddings, with the exception of MLP. CNNs had the highest scores across the board while LSTM had the lowest after the baseline, probably due to the small size of the dataset.

**Table 1:** F-scores of classifier-embedding type pairs. Pretrained: upper half, Ad hoc: lower half

|      | TOK   | TOK POS | LEM   | LEM POS | DEP   | TOK[10] | DEPC[10] |
|------|-------|---------|-------|---------|-------|---------|----------|
| SVM  | 0.481 | 0.483   | 0.484 | 0.483   | 0.48  | 0.481   | 0.497    |
| LSTM | 0.506 | 0.512   | 0.538 | 0.53    | 0.492 | 0.531   | 0.527    |
| CNN  | 0.754 | 0.712   | 0.757 | 0.702   | 0.654 | 0.751   | 0.749    |
| MLP  | 0.538 | 0.741   | 0.656 | 0.715   | 0.679 | 0.752   | 0.741    |
| SVM  | 0.793 | 0.791   | 0.784 | 0.788   | 0.568 |         |          |
| CNN  | 0.659 | 0.626   | 0.67  | 0.665   | 0.682 |         |          |
| MLP  | 0.796 | 0.787   | 0.786 | 0.783   | 0.594 |         |          |

Lemmas achieved the highest scores, being very slightly better than other embeddings with the exception of MLP, where Levy and Goldberg's [10] token embeddings and dependency context embeddings were clearly better. The differences in scores between our token embeddings and Levy and Goldberg's could be explained by the differences in the training data size as there is a noticeable difference in the vocabulary size of our embeddings versus theirs (40,000 vs 180,000). This could suggest that lemmatization can be effective as a technique in increasing the performance of embeddings for a cyberbullying detection task.

The POS embeddings did not score well compared to their simpler counterparts in most cases. The only exception being MLP where they showed a clear performance boost, with TOKPOS scoring especially high. One of the reasons for the generally lower performance could be related to the increased sparsity of the dataset due to adding POS information. This could be corrected by applying a larger dataset for training the word embeddings. The difference could also be in the forming of the embeddings themselves. Because of this, we are planning to conduct qualitative evaluation and manually inspect the embeddings more closely in the future.

Our dependency embeddings scored the lowest and were outperformed by all other types of embeddings in basically all cases. One of the reasons could be again related to the greatly increased sparsity of the dataset due to the added dependency information. In the future we plan to train the embeddings on a larger dataset and also conduct qualitative evaluation to study the differences between our implementation and Levy and Goldberg's [10].

According to the results, using dependency based embeddings does not offer any noticeable improvements on cyberbullying detection task. However, this needs to be confirmed using a larger training set for the embeddings. The task should also be evaluated using other cyberbullying datasets.

**(3) Comparison with *ad hoc* embeddings**

To see the effect of using pretrained word embeddings over *ad hoc* embeddings, we also ran the experiments with-out pretraining the embeddings. For the baseline SVM classifier, we used a tf-idf weighing scheme to produce a BoW language model of the evaluation (cyberbullying) dataset. For the neural network models, we trained the embeddings on the evaluation datasets themselves as part of the networks using Keras' embedding layer with random initial weights.

The results with *ad hoc* embeddings are shown in the lower half of Table 1. First thing to note is that SVM performed significantly better with the BoW language model instead of pretrained embeddings. The reason most likely is that SVM uses the embeddings as they are and no adjustment to the cyberbullying specific vocabulary is done during training of the classifier, whereas the BoW model was trained on the cyberbullying data itself and captures its concepts.

CNN's performance on the other hand was a significantly worse without using pretrained word embeddings. The difference in scores clearly shows why pretrained language models are popular, as the CNN model gains a noticeable boost from a very general dataset completely irrelevant to the target. This also shows the nature of the CNN model earlier observed by Kim [28], that CNNs greatly benefit from pretrained word embeddings. The same does not apply to MLP though as it seemed to perform slightly better without pretraining. With a larger pretraining dataset, the situation could be different.

## 5. CONCLUSIONS AND FUTURE WORK

In this paper we presented our research on linguistically-backed word embeddings, applied in cyberbullying detection. We showed that lemmatization can be used as an effective preprocessing method for increasing detection efficacy with pretrained word embeddings. From the experiment results it can also be seen that using dependency based embeddings does not increase performance of the classifiers.

We concluded that for SVM it is better to train the language model on the target data itself, whereas CNN benefits greatly from pretrained word embeddings. In the future, we are planning to do a qualitative evaluation on the different kinds of embeddings in order to analyze their effects more deeply. Also we are going to train the embeddings on larger datasets and measure the classification performance on other languages to confirm and further explore the results of this study.